\documentclass[conference]{IEEEtran}

\usepackage[utf8]{inputenc}
\usepackage{acro}
\usepackage{amsmath}
\usepackage{amsthm}
\usepackage{amssymb}
\usepackage{array}
\usepackage[bookmarks=true]{hyperref}
\usepackage[keeplastbox]{flushend}
\usepackage[a4paper, margin=24mm]{geometry}
\usepackage{graphicx}
\usepackage{multicol}
\usepackage{multirow}
\usepackage[numbers]{natbib}
\usepackage{pgfplots}
\usepackage{pgfplotstable}
\usepackage[separate-uncertainty=true]{siunitx}
\usepackage{tikz}
\usepackage{times}
\usepackage{url}

\usetikzlibrary{calc, decorations.markings}

\newtheorem{theorem}{Theorem}[section]
\newtheorem{lemma}[theorem]{Lemma}

\pgfplotsset{compat = newest}
\pgfplotsset{%
    every tick label/.append style = {font=\footnotesize},
    every axis label/.append style = {font=\small}
}

\DeclareAcronym{DoF}{short=DoF, long=degree of freedom, long-plural-form=degrees of freedom}
\DeclareAcronym{OTG}{short=OTG, long=online trajectory generation}
\DeclareAcronym{HRC}{short=HRC, long=human robot collaboration}

\begin{document}

\title{Jerk-limited Real-time Trajectory Generation \\ with Arbitrary Target States}


\author{\authorblockN{Lars Berscheid and Torsten Kröger}
\authorblockA{Karlsruhe Institute of Technology (KIT), Germany\\
\{lars.berscheid, torsten\}@kit.edu}
}

\maketitle

\begin{abstract}
We present \textit{Ruckig}, an algorithm for \ac{OTG} respecting third-order constraints and complete kinematic target states. Given any initial state of a system with multiple \acp{DoF}, Ruckig calculates a time-optimal trajectory to an arbitrary target state defined by its position, velocity, and acceleration limited by velocity, acceleration, and jerk constraints. The proposed algorithm and implementation allows three contributions: (1) To the best of our knowledge, we derive the first time-optimal \ac{OTG} algorithm for arbitrary, multi-dimensional target states, in particular including non-zero target acceleration. (2) This is the first open-source\footnote{\textit{Ruckig} is a C++ library published under the permissive MIT license at \url{https://github.com/pantor/ruckig}} prototype of time-optimal \ac{OTG} with limited jerk and complete time synchronization for multiple \acp{DoF}. (3) Ruckig allows for directional velocity and acceleration limits, enabling robots to better use their dynamical resources. We evaluate the robustness and real-time capability of the proposed algorithm on a test suite with over \num{1 000 000 000} random trajectories as well as in real-world applications.
\end{abstract}

\IEEEpeerreviewmaketitle

\section{Introduction}

Modern robots are supposed to operate in and manipulate their unknown and non-deterministic environments. It is therefore not sufficient to plan actions beforehand; instead robots need to be able to react to novel sensor input on the fly. Then, a new trajectory needs to be generated in a real-time manner, allowing the robot to adapt the task execution within the scope of its dynamical resources. The trajectory representation is particularly important, as it serves as an interface between the (more abstract) task planning and (lower level) motion planning. A common representation is \textit{waypoint-based}: The task execution is given as a single or list of waypoints with defined kinematic state. Then, \acf{OTG} will calculate a trajectory to the new waypoint target considering the robots constraints. Commonly, second-order (namely velocity and acceleration) constraints take the dynamical resources into account. However, third-order constraints (an additional jerk limit) are desirable to reduce mechanical stress, wear and tear, and the robots overall cost over lifetime. In fact, we find that modern industrial robots, e.g. by Franka Emika, monitor the jerk in the internal controller and terminate in case of acceleration discontinuities. \\

In this work, we propose a novel algorithm for \ac{OTG} named \textit{Ruckig} that is significantly simpler than related approaches. While \textit{guaranteeing} a solution even for systems with multiple \acp{DoF}, Ruckig enables three contributions to the field of \ac{OTG}: First, target waypoints can be defined not only by their position and velocity, but by their \textit{complete} kinematic state including acceleration. This improves the practical usability of \ac{OTG} in dynamic tasks. Second, Ruckig is the first open-source and freely available \ac{OTG} implementation with constrained jerk. Third, this work introduces directional velocity and acceleration limits. This makes it easier to exploit the full dynamic capabilities of the robot. Moreover, we argue that this is useful for human robot interaction: To ensure human safety, a velocity limit \textit{towards} the human needs to be met, while simultaneously moving away from the human to avoid possible contact. \\


In the following paper, we formalize the problem of waypoint-based \ac{OTG}, derive the proposed algorithm and share details about the implementation. We evaluate the robustness and real-time performance on a set of randomly generated trajectories. Finally, we show real-world applications highlighting the proposed contributions.

\begin{figure}[t]
    \centering

\pgfplotstableread{front-profile.txt}{\profile}
\begin{tikzpicture}
\begin{axis}[
    xmin = -0.1, xmax = 4.16,
    ymin = -1.3, ymax = 1.3,
    xtick distance = 1,
    ytick distance = 0.5,
    xticklabel=\empty,
    yticklabel=\empty,
    grid = major,
    minor tick num = 1,
    major grid style = {lightgray!50},
    width = 0.56\textwidth,
    height = 0.38\textwidth,
    legend style={
        at={(0.5, 1.03)},
        anchor=south,
        legend columns=-1
    },
]

    \draw[thin, black] (0.42, -2.0) -- (0.42, 2.0);
    \draw[thin, black] (0.91, -2.0) -- (0.91, 2.0);
    \draw[thin, black] (1.33, -2.0) -- (1.33, 2.0);
    \draw[thin, black] (2.38, -2.0) -- (2.38, 2.0);
    \draw[thin, black] (2.8, -2.0) -- (2.8, 2.0);
    \draw[thin, black] (3.81, -2.0) -- (3.81, 2.0);
    \draw[thin, black] (4.05, -2.0) -- (4.05, 2.0);
    
    \draw[teal, fill, fill opacity=0.018, dashed] (-1, 0.5) rectangle (5, 3);
    \draw[teal, fill, fill opacity=0.018, dashed] (-1, -0.5) rectangle (5, -3);
    \draw[orange, fill, fill opacity=0.05, dashed] (-1, 0.7) rectangle (5, 3);
    \draw[orange, fill, fill opacity=0.05, dashed] (-1, -0.7) rectangle (5, -3);
    \draw[red, fill, fill opacity=0.1, dashed] (-1, 1.2) rectangle (5, 3);
    \draw[red, fill, fill opacity=0.1, dashed] (-1, -1.2) rectangle (5, -3);
 
    \addplot[red!80!black, thick] table [x={t}, y={j}] {\profile};
    \addplot[teal, thick] table [x={t}, y={a}] {\profile};
    \addplot[orange, thick] table [x={t}, y={v}] {\profile};
    \addplot[blue!80!black, thick] table [x={t}, y={p}] {\profile};
    \legend{Jerk, Acceleration, Velocity, Position}
\end{axis}

    \node[] (t1) at (0.6, -0.4) {$t_1$};
    \node[] (t2) at (1.35, -0.4) {$t_2$};
    \node[] (t3) at (2.2, -0.4) {$t_3$};
    \node[] (t4) at (3.4, -0.4) {$t_4$};
    \node[] (t5) at (4.75, -0.4) {$t_5$};
    \node[] (t6) at (6.1, -0.4) {$t_6$};
    \node[] (t7) at (7.12, -0.4) {$t_7$};
\end{tikzpicture}
    \caption{A time-optimal profile of a single \ac{DoF} with initial velocity $v_0 \neq 0$ and target acceleration $a_f \neq 0$. The proposed algorithm is able to generate a time-synchronized trajectory for multiple \acp{DoF} with given velocity, acceleration and jerk constraints (dashed) in a limited number of operations (real-time capability).}
    \label{fig:profile}
\end{figure}
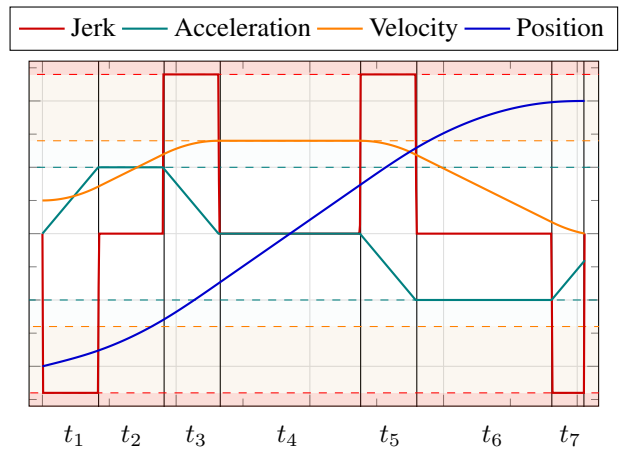

\section{Related Work}

As trajectory generation is an essential part within robotics, it has already seen decades of research of both \textit{offline} as well as \textit{online} (real-time) approaches.

\paragraph{Offline Trajectory Generation}

More than 50 years ago, Kahn et Roth were the first to use optimal linear control to generate near time-optimal trajectories \cite{kahn1969near}. Modern approaches commonly split the motion planning into: (1) a calculation of a geometric path, e.g. as a result of a task-specific (e.g. goal-finding or collision-avoidance) algorithm, and (2) a general time-parametrization of this path regarding the dynamic resources. Regarding this second step, Kunz et al. proposed a time-optimal time-parametrization with velocity and acceleration constraints \cite{kunz2012time}. In recent years, Pham et al. have improved the calculation time for second-order constraints significantly \cite{pham2014general}. Regarding jerk constraints, Lange et al. introduced an approach for path-following with constrained jerk \cite{lange2015path}. \\

\paragraph{Online Trajectory Generation}

For \ac{OTG}, the path and time-parametrization are oftentimes combined into a single step. This simplification is commonly done alongside a waypoint-based formulation of the problem. In contrast to most offline time-parametrization algorithms, \ac{OTG} only considers kinematic constraints of each \textit{independent} \ac{DoF}. This difference is also known as kinematic vs. dynamic time-optimality.

Macfarlane et al. introduced time-optimal and jerk-constrained \ac{OTG} for a single \ac{DoF} \cite{macfarlane2003jerk}. For multiple \acp{DoF}, Broquere et al. presented a jerk-limited \ac{OTG} however with zero initial acceleration \cite{broquere2008soft}.
Haschke et al. derived an algorithm to reach a given position with jerk-constraints from an arbitrary initial state with non-zero acceleration \cite{haschke2008line}.
Kröger et al. classified \ac{OTG} algorithms
\begin{table}[h]
    \centering
\begin{tabular}{l|>{\centering\arraybackslash}m{.12\textwidth}|>{\centering\arraybackslash}m{.12\textwidth}}
    \hline
    & Highest-order Non-zero Target & Highest-order Constraint \\
    \hline
    Type II & Velocity & Acceleration \\
    Type IV & Velocity & Jerk \\
    Type V & Acceleration & Jerk \\
    \hline
\end{tabular}
    \label{tab:type-classification}
\end{table}
depending on the highest derivative order of both the non-zero target state and constraint \cite{kroger2009online}. Subsequently, Kröger introduced \textit{Reflexxes}, an open-source \textit{Type II} and proprietary \textit{Type IV} implementation \cite{kroger2011opening}. The latter allows to reach waypoints with non-zero velocity from any arbitrary state.
In this context, the proposed \textit{Ruckig} algorithm and library is - to the best of our knowledge - the first \textit{Type V} and the first open-source \textit{Type IV} implementation. \\

While considering complete third-order states, Ahn et al. presented a non-time-optimal algorithm that does not consider velocity, acceleration, or jerk limits directly \cite{ahn2004arbitrary}. In the field of quadrotor flight, Beul et al. introduced \textit{opt$\_$control}, an open-source jerk-constrained \ac{OTG} algorithm however without considering complete time synchronization for arbitrary input states \cite{beul2016analytical}. Moreover, there is significant interest in the field of safety and \ac{HRC}. Most prominent, Haddadin et al. investigated reaction motions in the context of \ac{OTG} for safe human-robot interaction \cite{haddadin2008collision}.

\section{Problem Definition}

Let $\vec{x}$ be the state of a kinematic particle with $N$ \acp{DoF}. The state $x_i(t)$ of \ac{DoF} $i \in \lbrace 1, \dots, N \rbrace$ at time $t$ is defined by the position $p_i$ and its partial derivatives of up to \textit{third order}
\begin{align*}
    x_i = \left( p_i, \, v_i := \frac{\partial p_i}{\partial t}, \, a_i := \frac{\partial^2 p_i}{\partial t^2}, \, j_i := \frac{\partial^3 p_i}{\partial t^3} \right)
\end{align*}
named velocity $v_i$, acceleration $a_i$, and jerk $j_i$. We consider the kinematic time-optimality, otherwise the total instead of partial derivatives would be required. Given an initial state $\vec{x}_0$ and a target (final) state $\vec{x}_f$, we seek the time-optimal trajectory $\vec{x}^{*}(t)$ defined by
\begin{align*}
    \vec{x}^{*}(t) = \text{arg}\min_{\vec{x}(t)} T_f, \quad \vec{x}(0) = \vec{x}_0, \quad \vec{x}(T_f) = \vec{x}_f
\end{align*}
satisfying the velocity, acceleration and jerk constraints
\begin{align*}
    v_{i,min} ~ \leq ~ &v_i(t) ~ \leq ~ v_{i,max} \\
    a_{i,min} ~ \leq ~ &a_i(t) ~ \leq ~ a_{i,max} \\
    j_{i,min} ~ \leq ~ &j_i(t) ~ \leq ~ j_{i,max}
\end{align*}
for all times $t \in [0, T_f]$ and \acp{DoF} $i$. $T_f$ is called the trajectory duration. If no vector-notation is given furthermore, we calculate each \ac{DoF} independently. Note that we consider both the initial as well as target state to be \textit{complete} with possibly all derivatives non-zero. For simplicity, we assume $j_{min} = -j_{max}$ furthermore, but keep the directional acceleration and velocity limits. Moreover, not every kinematic target state is physically possible. We define an upper bound of the allowed target acceleration by
\begin{align}
    a_{f} \leq \sqrt{2 j_{max} \max \left( \vert v_{max} - v_{f} \vert, \vert v_{min} - v_{f} | \right)},
    \label{eq:maximum-acceleration}
\end{align}
because an acceleration target requires a minimum velocity interval to be reached with constrained jerk.

\section{Algorithm}

We divide the \ac{OTG} problem into six subsequent steps. Following a short overview, each step is explained in detail in its own subsection.
\begin{LaTeXdescription}
    \item[A.] An optional \textbf{brake pre-trajectory} is calculated if the initial state $x_0$ exceeds or will inevitably exceed the kinematic limits $v_{min}$, $v_{max}$, $a_{max}$, or $a_{min}$. In this case, recovering to a safe kinematic state is the most urgent task.
    \item[B.] In \textbf{Step 1: Extremal times}, all possible profiles that utilize the full dynamic resources of the robot are calculated for each \ac{DoF} $i$ independently. We call this the set of \textit{valid extremal profiles}. The duration of the fastest profile is called $T_{i,min}$.
    \item[C.] The target $\vec{x}_f$ should be reached at the same time point $T_f$ by each \ac{DoF} $i$. Therefore, some \acp{DoF} might need to slow down. In general, not every trajectory duration $T > \max_i (T_{i,min})$ is possible, as there might be a limited number of \textbf{blocked intervals} for the duration. We derive these intervals based on the set of valid extremal profiles.
    \item[D.] The \textbf{minimum trajectory duration} $T_f$ is the fastest duration that is not blocked by any \ac{DoF}. This duration corresponds to a limiting profile as well as a limiting \ac{DoF} and is included in the set of valid profiles.
    \item[E.] In \textbf{Step 2: Time synchronization}, we calculate a profile for every \ac{DoF} that reaches its target $x_f$ at the given trajectory duration $T_f$.
    \item[F.] Finally, the \textbf{new state} at a given time $t$ on the trajectory can be calculated.
\end{LaTeXdescription}

A time-optimal trajectory will be limited by a single \ac{DoF} $l$ which uses its entire dynamical resources at all times $t$. Therefore, this \ac{DoF} will use a \textit{bang-bang}-like jerk profile with $j_l(t) \in \lbrace -j_{max}, 0, j_{max} \rbrace$. We call such profiles \textit{extremal}. In particular, the total duration $T$ of a profile can be changed by its underlying jerk profile in infinitesimal steps. If and only if the profile is extremal, the duration is bounded on one side.

More generally, we formulate each trajectory as a sequence of constant jerk values $j_k = s_k j_f$ with corresponding non-negative time steps $t_k \geq 0$. Let $s_k \in \lbrace -1, 0, 1 \rbrace$ be the jerk sign and $j_f > 0$ the jerk value constant throughout the profile. An extremal profile with third-order constraints results in a linear acceleration, quadratic velocity, and in a cubic polynomial for the position. Fig.~\ref{fig:profile} shows an illustrative example of an extremal profile with seven steps $k$.

\subsection{Brake Pre-trajectory}

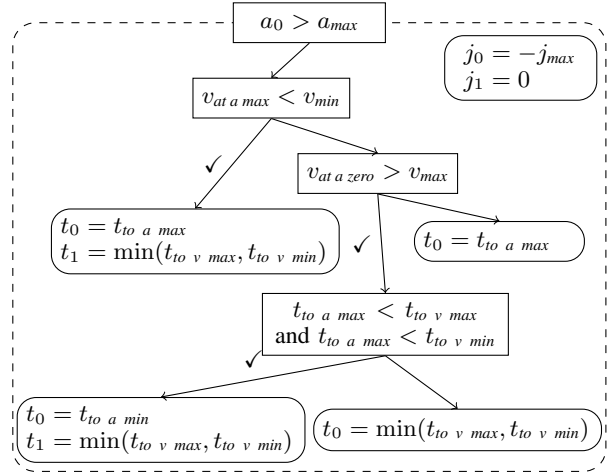
\begin{figure}[t]
    \centering
\begin{tikzpicture}
    \tikzstyle{block1} = [draw, text centered, minimum width=2cm, minimum height=1.5em, font=\small, fill=white]
    \tikzstyle{block2} = [draw, text centered, minimum width=2cm, minimum height=1.5em, font=\small, rounded corners=0.25cm, fill=white, align=left]
    
    \draw[dashed, rounded corners=0.3cm] (-3.9, 0.0) rectangle (3.9, -5.95);
    
    \node[block1] (a1) {$a_0 > a_{\textit{max}}$};
    
    \node[block2, text width=1.5cm] (j1) at (2.78, -0.6) {$j_0 = -j_{\textit{max}}$\\ $j_1 = 0$};
    
    \node[block1] (v1)  at (-0.5, -1.0) {$v_{\textit{at a max}} < v_{\textit{min}}$};
    
    \node[block1] (v2) at (0.9, -2.0) {$v_{\textit{at a zero}} > v_{\textit{max}}$};
    
    \node[block2, text width=3.55cm] (s2) at (-1.5, -2.9) {$t_0 = t_{\textit{to a max}}$ \\ $t_1 = \min(t_{\textit{to v max}}, t_{\textit{to v min}})$};
    
    \node[block2, text width=2cm] (s1) at (2.5, -2.9) {$t_0 = t_{\textit{to a max}}$};
    
    \node[block1, text width=3.0cm] (t1) at (1.0, -4.0) {$t_{\textit{to a max}} < t_{\textit{to v max}}$\\ and $t_{\textit{to a max}} < t_{\textit{to v min}}$};
    
    \node[block2, text width=3.55cm] (s3) at (-1.95, -5.4) {$t_0 = t_{\textit{to a min}}$ \\ $t_1 = \min(t_{\textit{to v max}}, t_{\textit{to v min}})$};
    
    \node[block2, text width=3.55cm] (s4) at (1.95, -5.4) {$t_0 = \min(t_{\textit{to v max}}, t_{\textit{to v min}})$};
    
    \draw[->] (a1.south) -- (v1.north);
    \draw[->] (v1.south) -- (s2.north) node[midway, left] {\small $\checkmark$};
    \draw[->] (v1.south) -- (v2.north);
    \draw[->] (v2.south) -- (t1.north) node[midway, left] {\small $\checkmark$};
    \draw[->] (v2.south) -- (s1.north);
    \draw[->] (t1.south) -- (s3.north) node[midway, above left] {\small $\checkmark$};
    \draw[->] (t1.south) -- (s4.north);
\end{tikzpicture}
    \caption{A part of the decision tree (given $a_0 > a_{max}$) for calculating an optional brake pre-trajectory. If required, we calculate the time-optimal profile to transfer the system to a safe kinematic state. The pre-trajectory is determined by up to two steps of constant jerk $j_0$ and $j_1$ and their respective duration $t_0$ and $t_1$.}
    \label{fig:brake-trajectory}
\end{figure}

If the initial state $x_0$ exceeds or will exceed the acceleration or velocity limits, a so-called pre-trajectory is introduced to brake the system below its respective limits. Due to limited jerk, cases exists that will inevitably brake the velocity constraints at a later point in time. Let $T_{ib}$ be the duration of the brake pre-trajectory or zero if none is required. In comparison to the rest of the algorithm, this step works in the velocity domain ignoring any position values. We introduce a decision tree (Fig.~\ref{fig:brake-trajectory})
depending on $a_0$, $v_0$, $a_{max}$, $a_{min}$, $v_{max}$, and $v_{min}$ that calculates the fastest profile to reach the limits. It can be seen by distinction of cases that a resulting profile includes up to two time-steps $t_{ib0}$ and $t_{ib1}$ with corresponding jerk $j_{ib0} \in \lbrace -j_{max}, j_{max} \rbrace$ and $j_{ib1} = 0$. The second step with zero jerk might be necessary, as no \textit{new} constraints should get broken.

\subsection{Step 1: Extremal Times}

We want to find all extremal profiles that reach the target state $x_f$ for all \acp{DoF} independently.

\begin{lemma}
A velocity limit might only be reached once in an extremal profile.
\end{lemma}
At a velocity limit, the profile has zero acceleration and zero jerk. The profile can always be decelerated by reducing the velocity plateau. We show by contradiction: If two velocity limits would be in the same direction, the profile could be accelerated by removing the intermediate deceleration and extending the maximal velocity. If the velocity limits would be in opposite directions, the profile could be accelerated by removing the distance traveled from the shorter direction from the other one. As the duration can be shortened and extended, it cannot be an extremal profile.

\begin{lemma}
There are only up to two acceleration limits in an extremal profile.
\end{lemma}
Otherwise, a third acceleration peak exists resulting in one direction reached at least two times. Then, the profile could be speed up by shifting the acceleration from the later reached peak to the prior one. The profile could be decelerated with the inverse approach. \\

Therefore, only up to three limits can be reached in total (as shown in Fig.~\ref{fig:profile}): First, an acceleration limit called \texttt{ACC0}, second a velocity limit called \texttt{VEL}, and third an acceleration limit called \texttt{ACC1}. Introducing optional steps of constant jerk before, after, and between the limits lead to a maximal number of seven steps $k$ with jerk $j_k = s_k j_f$ and corresponding duration $t_k$ (Table~\ref{tab:jerk-steps}). We denote the sign of the non-zero jerk $s_k$ as either $\uparrow = +1$ or $\downarrow = -1$. A redundant step is encoded with zero duration $t_k = 0$. 
\begin{table}[b]
	\centering
	\caption{Steps of Constant Jerk of an Extremal Profile}
	\begin{tabular}{c|c|c}
	\hline
	Step $k$ & Jerk Sign $s_k$ & Limit \\
	\hline
	$t_1$ & $\uparrow$ \textit{or} $\downarrow$ & - \\
	$t_2$ & 0 & \texttt{ACC0} \\
	$t_3$ & $\uparrow$ \textit{or} $\downarrow$ & - \\
	$t_4$ & 0 & \texttt{VEL} \\
	$t_5$ & $\uparrow$ \textit{or} $\downarrow$ & - \\
	$t_6$ & 0 & \texttt{ACC1} \\
	$t_7$ & $\uparrow$ \textit{or} $\downarrow$ & - \\
	\hline
	\end{tabular}
	\label{tab:jerk-steps}
\end{table}

\begin{table*}[ht]
	\centering
	\caption{Time-Optimal Profile Types for the \texttt{UP} Direction.}
	\vspace{1mm}
	\begin{tabular}{|c|c|l|l|l|l|l|}
	\hline
	Step & Jerk Profile & Limits & Condition I & Condition II  & Condition III & Condition IV \\
	\hline
	\hline
	\multirow{12}{*}{Step 1 + 2} & \multirow{8}{*}{$\uparrow\downarrow\downarrow\uparrow$} & \texttt{ACC0} \texttt{VEL} \texttt{ACC1} & $a_{1} = a_{max}$ & $v_{3} = v_{max}$ & $a_{5} = a_{min}$ & $a_3 = 0$ \\
	& & \texttt{ACC0} \texttt{VEL} & $a_{1} = a_{max}$ & $v_{3} = v_{max}$ & $t_{6} = 0$ & $a_3 = 0$ \\
	& & \texttt{VEL} \texttt{ACC1} & $t_{2} = 0$ & $v_{3} = v_{max}$ & $a_{5} = a_{min}$ & $a_3 = 0$ \\
	& & \texttt{VEL} & $t_{2} = 0$ & $v_{3} = v_{max}$ & $t_{6} = 0$ & $a_3 = 0$ \\
	& & \texttt{ACC0} \texttt{ACC1} & $a_{1} = a_{max}$ & $t_{4} = 0$ & $a_{5} = a_{min}$ & $t_5 = 0$ \\
	& & \texttt{ACC0} & $a_{1} = a_{max}$ & $t_{4} = 0$ & $t_{6} = 0$ & $t_5 = 0$ \\
	& & \texttt{ACC1} & $t_{2} = 0$ & $t_{4} = 0$ & $a_{5} = a_{min}$ & $t_5 = 0$ \\
	& & \texttt{NONE} & $t_{2} = 0$ & $t_{4} = 0$ & $t_{6} = 0$ & $t_5 = 0$ \\
	\cline{2-7}
	& \multirow{4}{*}{$\uparrow\downarrow\uparrow\downarrow$} & \texttt{ACC0} \texttt{ACC1} & $a_{1} = a_{max}$ & $t_{4} = 0$ & $a_{5} = a_{max}$ & $t_7 = 0$ \\
	& & \texttt{ACC0} & $a_{1} = a_{max}$ & $t_{4} = 0$ & $t_{6} = 0$ & $t_7 = 0$ \\
	& & \texttt{ACC1} & $t_{2} = 0$ & $t_{4} = 0$ & $a_{5} = a_{max}$ & $t_7 = 0$ \\
	& & \texttt{NONE} & $t_{2} = 0$ & $t_{4} = 0$ & $t_{6} = 0$ & $t_7 = 0$ \\
	\hline
	\hline
	\multirow{4}{*}{Step 2} & \multirow{4}{*}{$\uparrow\downarrow\uparrow\downarrow$} & \texttt{ACC0} \texttt{VEL} \texttt{ACC1} & $a_{1} = a_{max}$ & $v_{3} = v_{max}$ & $a_{5} = a_{max}$ & $t_7 = 0$ \\
	& & \texttt{ACC0} \texttt{VEL} & $a_{1} = a_{max}$ & $v_{3} = v_{max}$ & $t_{6} = 0$ & $t_7 = 0$ \\
	& & \texttt{VEL} \texttt{ACC1} & $t_{2} = 0$ & $v_{3} = v_{max}$ & $a_{5} = a_{max}$ & $t_7 = 0$ \\
	& & \texttt{VEL} & $t_{2} = 0$ & $v_{3} = v_{max}$ & $t_{6} = 0$ & $t_7 = 0$ \\
	\hline
	\end{tabular}
	\label{tab:profile-types}
\end{table*}

Four non-zero jerk steps result in \num{16} possible combinations. However, only four unique profiles meet the kinematic constraints and are non-redundant: $\uparrow\downarrow\downarrow\uparrow$, $\uparrow\downarrow\uparrow\downarrow$, $\downarrow\uparrow\uparrow\downarrow$, and $\downarrow\uparrow\downarrow\uparrow$. As the overall problem is invariant to a sign change in $j_{max}$ and exchanging $v_{min} \leftrightarrow v_{max}$ and $a_{min} \leftrightarrow a_{max}$, the set of distinct jerk profiles can be simplified further to $\uparrow\downarrow\downarrow\uparrow$ and $\uparrow\downarrow\uparrow\downarrow$ profiles. Then, the first jerk sign corresponds to the \texttt{UP} or \texttt{DOWN} direction.

The final list of profile types include every combination of the above three limits and final two jerk profile types. Table~\ref{tab:profile-types} lists all 16 distinct profile types for a single direction. In step 1, only \num{12} profiles are possible as a $\uparrow\downarrow\uparrow\downarrow$ profile with a positive acceleration after the velocity limit is not valid.

Mathematically, each profile type maps the initial state $x_0$, the target state $x_f$, and the given limits
\begin{align*}
    &\mathcal{S}1: (p_0, p_f, v_0, v_f, a_0, a_f, v_{max}, a_{max}, a_{min}, j_{max}) \\
    &\quad \mapsto (t_1, t_2, t_3, t_4, t_5, t_6, t_7)
\end{align*}
to corresponding times $t_1$ to $t_7$. Given \num{3} equations for position, velocity and acceleration and \num{7} variables, \num{4} additional conditions need to be introduced. Three conditions are set by their limits or a zero step duration. The final condition is either set to $a_3 = 0$ for constant velocity, $t_5 = 0$ for fusing the centering steps $\uparrow\downarrow\downarrow\uparrow$ if $t_4 = 0$, or $t_7 = 0$ to reach time-optimality for the $\uparrow\downarrow\uparrow\downarrow$ profile. Note that profiles might have multiple solutions. Most profiles are analytically solvable.
For some profiles however, roots of up to sixth-order polynomials need to be found. Here, we make use of a safe Newton root-finding algorithm: Given an isolated root of a polynomial in an interval, a Newton method ensures quadratic convergence on average. Using a bisection method as a fallback strategy, an upper bound of the number of iterations for a given tolerance can be specified. This is required to ensure real-time capability. The initial interval is found by solving the second derivative of the sixth-order polynomial analytically and check if a root exists between two extrema. This step is repeated for the first derivative, leading to intervals with isolated roots for the polynomial itself. \\

We calculate the numeric times $t_k$ for all \num{24} possible profile types, as given in Table~\ref{tab:profile-types} per direction. However, not all solutions are physically reasonable or within the kinematic limits of the system. First, we check that every time step $1 \leq k \leq 7$ is non-negative
\begin{align*}
    t_k ~ &\geq ~ 0.
\end{align*}
Then, we integrate position, velocity and acceleration
\begin{align}
    a_{k+1} &= a_k + s_k j_k t_k, \label{eq:acc-integration} \\
    v_{k+1} &= v_k + a_k t_k + \frac{s_k j_k}{2} t_k^2, \label{eq:vel-integration} \\
    p_{k+1} &= p_k + v_k t_k + \frac{a_k}{2} t_k^2 + \frac{s_k j_k}{6} t_k^3
    \label{eq:pos-integration}
\end{align}
for each time step. We check the acceleration limits via
\begin{align*}
     a_{min} < \lbrace \, a_1, a_3, a_5 \, \rbrace &< a_{max}
\end{align*}
and the velocity limits via
\begin{align*}
    v_{min} < v_k - \frac{a_k^2}{2 s_k j_k} &< v_{max} \quad \text{if} \quad a_k \cdot a_{k+1} \leq 0, \\
    v_{min} < v_k &< v_{max} \quad \text{if} \quad a_k = 0.
\end{align*}
If the profile passes all checks, it is added to the set of \textit{valid extremal profiles}. The total duration is given by
\begin{align}
	T = \sum_{k=1}^{7} t_k.
\end{align}
We find the fastest profile and its duration $T_{i,min}$ for each \ac{DoF} $i$ easily by comparing.

\subsection{Blocked Duration Intervals}

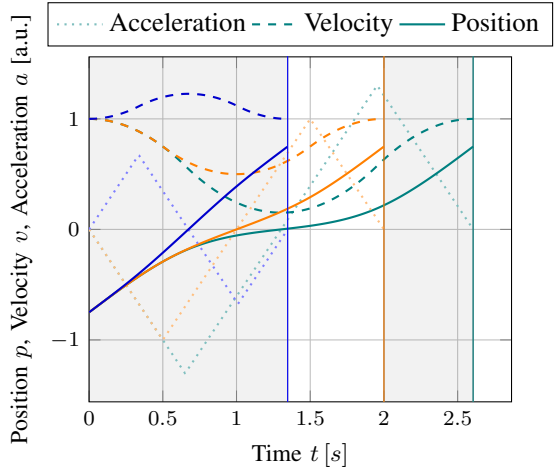
\begin{figure}[ht]
    \centering

\pgfplotstableread{blocked-interval-1.txt}{\intervala}
\pgfplotstableread{blocked-interval-2.txt}{\intervalb}
\pgfplotstableread{blocked-interval-3.txt}{\intervalc}
\begin{tikzpicture}
\begin{axis}[
    grid=both,
    height=0.38\textwidth,
    ylabel={Position $p$, Velocity $v$, Acceleration $a$ [a.u.]},
    xlabel={Time $t \, [s]$},
    xmin=0,
    legend style={
        at={(0.5, 1.03)},
        anchor=south,
        legend columns=-1
    },
]

    \draw[black!50, fill, fill opacity=0.1] (-2.0, 2.0) rectangle (1.347, -2.0);
    \draw[black!50, fill, fill opacity=0.1] (2.0, 2.0) rectangle (2.605, -2.0);

    \addplot[teal!50, thick, dotted] table [x={t}, y={a}] {\intervalc};
    \addplot[teal, thick, dashed] table [x={t}, y={v}] {\intervalc};
    \addplot[teal, thick] table [x={t}, y={p}] {\intervalc};

    \addplot[orange!50, thick, dotted] table [x={t}, y={a}] {\intervalb};
    \addplot[orange, thick, dashed] table [x={t}, y={v}] {\intervalb};
    \addplot[orange, thick] table [x={t}, y={p}] {\intervalb};
    
    \addplot[blue!50, thick, dotted] table [x={t}, y={a}] {\intervala};
    \addplot[blue!80!black, thick, dashed] table [x={t}, y={v}] {\intervala};
    \addplot[blue!80!black, thick] table [x={t}, y={p}] {\intervala};
    
    \legend{Acceleration, Velocity, Position}
    
    \draw[thin, teal] (2.605, 2.0) -- +(0, -10.0);
    \draw[thin, orange] (2.0, 2.0) -- +(0, -10.0);
    \draw[thin, blue] (1.347, 2.0) -- +(0, -10.0);
\end{axis}
\end{tikzpicture}
    \caption{Example of a single blocked interval: Given $p_0 = -0.75$, $p_f = 0.75$, and $v_0 = v_f = 1.0$. No trajectory is physically possible with a duration below $t_{min} = 1.35$ \textit{and} between $t \in (2.0, 2.6)$.}
    \label{fig:example-blocked-interval}
\end{figure}

Given the set of valid extremal profiles, we want to find all possible duration $T_{i} > T_{i,min}$. In general, a number of \textit{blocked intervals} $(T_{1\alpha,start}, T_{1\alpha})$ might exist, in which a \ac{DoF} cannot reach the target with a duration within the interval. Fig.~\ref{fig:example-blocked-interval} illustrates an example for a single blocked interval.

\begin{lemma}
For a third-order target state, up to two blocked intervals might exist.
\end{lemma}
Here, we refer to the work of Kröger et al. \cite{kroger2009online}. In particular, the maximal number of blocked intervals depend on the target velocity and acceleration being non-zero. In our case, we denote the two possible blocked intervals as $\alpha$ and $\beta$.
\begin{table}[h]
	\centering
	\begin{tabular}{c|c|c}
	\hline
	Velocity & Acceleration & Max. Number of Blocked Intervals \\
	\hline
	$v_f = 0$ & $a_f = 0$ & 0 \\
	$v_f \neq 0$ & $a_f = 0$ & 1 \\
	- & \textbf{$a_f \neq 0$} & \textbf{2} \\
	\hline
	\end{tabular}
	\label{tab:blocked-intervals}
\end{table}

Furthermore, we want to clarify the relationship between blocked intervals and extremal profiles.
\begin{lemma}
A blocked interval is between two extremal profiles and each valid extremal profile corresponds to an interval boundary.
\end{lemma}
For all but extremal profiles, the duration can be adapted infinitesimally by changing the jerk $j_f$ or introducing a velocity or acceleration plateau. As the duration is constrained (to one side) for a boundary profile, it must be extremal and vice versa.
\begin{lemma}
A blocked interval can only exist between two neighboring profiles (regarding their duration).
\end{lemma}
Otherwise there would be a valid profile within a blocked interval. \\

Given up to two intervals, the set of valid extremal profiles must include \textit{exactly} $1$, $3$, or $5$ profiles. Given a sorted list of the profile duration, the profiles are mapped to blocked intervals as follows:
\begin{LaTeXdescription}
    \item[One profile] results in no blocked intervals.
    \item[Three profiles] lead to a single blocked interval $\alpha$. The interval is between the second and third profile.
    \item[Five profiles] correspond to two blocked intervals $\alpha$ and $\beta$. The first interval is between the second and third profile, the second one between the fourth and fifth profile.
    \item[Else] a failure of the algorithm would have occurred. In particular, the implementation has to deal with edge cases where different profile types merge.
\end{LaTeXdescription}

\subsection{Minimum Duration}

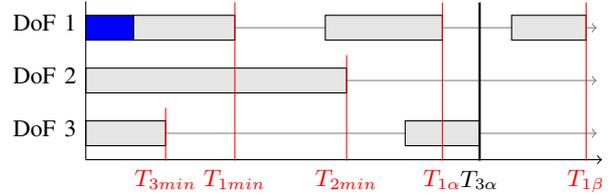
\begin{figure}[ht]
    \centering
\begin{tikzpicture}[font=\small, scale=0.7]
    \def\bary{1.0}
    \def\barheight{0.47}

    \draw[thin, gray, ->] (0, 2*\bary) -- (9.6, 2*\bary);
    \draw[thin, gray, ->] (0, \bary) -- (9.6, \bary);
    \draw[thin, gray, ->] (0, 0) -- (9.6, 0);

    \draw[fill=blue] (0.0, 2*\bary - \barheight/2) node [left, anchor=south east] {DoF 1} rectangle ++(0.9, \barheight);
	\draw[fill=gray!20] (0.9, 2*\bary - \barheight/2) rectangle ++(1.9, \barheight);
	\draw[fill=gray!20] (4.5, 2*\bary - \barheight/2) rectangle ++(2.2, \barheight);
	\draw[fill=gray!20] (8, 2*\bary - \barheight/2) rectangle ++(1.4, \barheight);
	
	\draw[fill=gray!20] (0, \bary - \barheight/2) node [left, anchor=south east] {DoF 2}  rectangle ++(4.9, \barheight);
	
	\draw[fill=gray!20] (0, -\barheight/2) node [left, anchor=south east] {DoF 3} rectangle ++(1.5, \barheight) ;
	\draw[fill=gray!20] (6, -\barheight/2) rectangle ++(1.4, \barheight);
	
	\draw (0, -0.5) -- ++(0, 3*\bary);
	\draw[->] (0, -0.5) -- ++(9.7, 0);
	
	\draw[thin, red] (1.5, -0.52) -- ++(0, \bary) node [below, at start] {$T_{3min}$};
	\draw[thin, red] (2.8, -0.52) -- ++(0, 3*\bary) node [below, at start] {$T_{1min}$};
	\draw[thin, red] (4.9, -0.52) -- ++(0, 2*\bary) node [below, at start] {$T_{2min}$};
	
	\draw[thin, red] (6.7, -0.52) -- ++(0, 3*\bary) node [below, at start] {$T_{1\alpha}$};
	\draw[thick, black] (7.4, -0.52) -- ++(0, 3*\bary) node [below, at start] {$T_{3\alpha}$};
	
	\draw[thin, red] (9.4, -0.52) -- ++(0, 3*\bary) node [below, at start] {$T_{1\beta}$};
\end{tikzpicture}

    \caption{Example of finding the minimum non-blocked duration of multiple \acp{DoF}. Shown are the blocked intervals (gray), possible minimum duration (red), possible braking pre-trajectories (blue), and the final minimum duration (black).}
    \label{fig:min-duration}
\end{figure}

Given the blocked duration intervals for each \ac{DoF}, we want to find the minimum duration that is possible for \textit{all} \acp{DoF}. Fig.~\ref{fig:min-duration} shows an exemplary illustration of this problem.

The minimum duration $T_f$ needs to be either time-optimal for a single \ac{DoF} or correspond to the right boundary of a blocked interval:
\begin{align*}
    T_f ~ \in ~ &\lbrace ~ \forall i \in \lbrace 0, \dots, N \rbrace: \\
    &\quad T_{ib} + T_{i,min}, \\
    &\quad T_{ib} + T_{i\alpha} \quad \text{ if $\alpha$ exists }, \\
    &\quad T_{ib} + T_{i\beta} \quad \text{ if $\beta$ exists } \rbrace
\end{align*}
The duration of a possible braking pre-trajectory needs to be added. Then, up to $3N$ possible duration are sorted and evaluated in ascending order. The first duration that is not blocked in any \ac{DoF} is the \textit{final} trajectory duration $T_f$. The \ac{DoF} $l$ that corresponds to the resulting duration is called the \textit{limiting} \ac{DoF}.

\subsection{Step 2: Time Synchronization}

Given the trajectory duration $T_f$, let $T_{ip} = T_f - T_{ib}$ be the duration of the profile without possible braking. Expect for the limiting \ac{DoF} $l$ (where we can reuse the calculated profile from the prior step), we need to find trajectories of corresponding duration $T_{ip}$. Therefore, step 2 maps the duration $T_f$, the initial state $x_0$, the target state $x_f$ and the given limits
\begin{align*}
    &\mathcal{S}2: (T_{p}, p_0, p_f, v_0, v_f, a_0, a_f, v_{max}, a_{max}, a_{min}, j_{max}) \\
    &\quad \mapsto (t_1, t_2, t_3, t_4, t_5, t_6, t_7, j_f)
\end{align*}  
to corresponding times $t_k$ and the final jerk constant $j_f$. In comparison the extremal profiles, we adapt the duration by changing the velocity plateau \texttt{VEL} to $v_{plat}$ with $v_{min} < v_{plat} < v_{max}$ or by reducing the jerk $|j_f| < j_{max}$. With a velocity plateau below its limit, the $\uparrow\downarrow\uparrow\downarrow$ profile gets possible for all profile types with \texttt{VEL} limit, leading to the full \num{16} possible profiles for each direction (Table.~\ref{tab:profile-types}). Here, we check all \num{32} profiles similarly to step 1, but return after the first valid profile is found.

\subsection{New State}

So far, \textit{Ruckig} has calculated the step duration $t_k$ and corresponding jerk signs $s_k$, the jerk value $j_f$, and a possible two-step brake pre-trajectory $t_{ib}$ and $j_{b}$ for each \ac{DoF} $i$. For each step $k \leq 7$, we integrate the acceleration $a_{k+1}$, velocity $v_{k+1}$ and position $p_{k+1}$ of the final profile according to (\ref{eq:acc-integration}), (\ref{eq:vel-integration}), and (\ref{eq:pos-integration}). Then, we can calculate the state at a given time $t$ by finding the last index $s$ that fulfills
\begin{align*}
    \sum_{i=1}^s t_i \leq t
\end{align*}
and integrating from index $s$ for time $t - t_s$ starting from the kinematic state $p_s$, $v_s$, and $a_s$. If a brake trajectory exists, we apply the same principle to this pre-trajectory. Usually, only this final integration of the new state will be repeated every control cycle of the robotic system. If and only if the input parameters change, the whole trajectory needs to be recalculated.

\section{Experimental Results}

\textit{Ruckig} is available under the permissive MIT license at \url{https://github.com/pantor/ruckig}. It is implemented as a C++17 library without further dependencies. Symbolic equations were solved ahead of time using \textit{Wolfram Mathematica}; the corresponding notebooks are included in the repository. The generated equations were exported as C/C++. To keep the implementation simple, we preferred to export a polynomial which roots correspond to the profile solution. This form is then solved by our own C++ polynomial root solver. A Python wrapper using \textit{pybind11} for offline trajectory generation is available. Moreover, we've implemented a velocity-control interface which calculates time-optimal trajectories ignoring the current position, the target position and velocity limits. Due to its simplicity, we focus only on the complete position-control interface in this paper.

\subsection{Robustness}

To evaluate the robustness and numerical stability of the proposed algorithm, we generated a test suite of over \num{1 000 000 000} random trajectories with up to 7 \acp{DoF} each. The input parameters are drawn from
\begin{align*}
    p_0, \, p_f &\sim \mathcal{N}(\sigma = 4.0) \\
    v_0, \, v_f, \, a_0, \, a_f &\sim \mathcal{N}(\sigma = 0.8) \\
    v_{max}, \, a_{max}, \, j_{max} &\sim \mathcal{G}(2.0, 2.0) + \num{0.05}
\end{align*}
with the Normal distribution $\mathcal{N}$, the gamma distribution $\mathcal{G}$ and a minimum limit of \num{5e-2}. We skip cases that violate the target acceleration requirement (\ref{eq:maximum-acceleration}). Then, we define a successful calculation if the maximal deviations $\Delta$ between the result and the target state of
\begin{align*}
    | \Delta_p | < 10^{-8}, \quad | \Delta_v | < 10^{-8}, \quad | \Delta_a | < 10^{-12}
\end{align*}
are met. Here, we achieve a robustness of \SI{100}{\%}. However, Ruckig is quite sensitive to long trajectory duration. During integration, the numerical error will propagate with $\Delta_p = T^2 \Delta_a$. With the precision of a double type for $a$ and the required position accuracy of $p$, this results in a maximal trajectory duration of
\begin{align*}
    T &= \sqrt{\frac{\Delta_p}{\Delta_a}} \approx \sqrt{\frac{\num{1e-8}}{\num{2e-16}}} = \num{7.1e3}
\end{align*}
that fulfills the above numerical error. In SI-units, this corresponds to an upper limit of around \SI{16}{min}. If cases above this maximum valid trajectory duration are ignored, Ruckig achieves a robustness of \SI{100}{\%} even without any minimum limit value. Note that the input parameters are invariant to the unit of distance, so the input can be scaled without loss of generality. \\

We compare the duration of trajectories generated by Ruckig and Reflexxes Type IV for the above input distribution with $a_f = 0$. We find that both duration of every trajectory within our test suite are within a deviation of $| \Delta_t | < 10^{-6}$, supporting the claim of time-optimality of each other.

\subsection{Calculation Duration}

As an online trajectory generator, Ruckig is real-time critical and must output the next state within one control cycle of the robot. Typical control cycles range between \SI{0.5}{ms} and \SI{5}{ms}. Following measurements were done on an Intel i7-8700K CPU 3.70GHz 6-core CPU using a single thread with PREEMPT-Linux. The test suite for benchmarking reuses the above input distribution.

\begin{table}[ht]
	\centering
	\caption{Calculation Performance for 7 \acp{DoF}}
	\vspace{1mm}
	\begin{tabular}{|lr|c|c|}
	\hline
	& & Mean [\si{\mu s}] & Worst [\si{\mu s}] \\
	\hline
	\hline
	Ruckig (ours) & $(a_f \neq 0)$ & \num{19.8 \pm 0.2} & \num{123 \pm 13} \\
	Reflexxes Type IV & $(a_f = 0)$ & \num{38.4 \pm 0.4} & \num{155 \pm 35} \\
	opt$\_$control & $(a_f \neq 0)$ & \num{727 \pm 7} & \num{3203 \pm 504} \\
	\hline
	\end{tabular}
	\label{tab:calculation-duration}
\end{table}

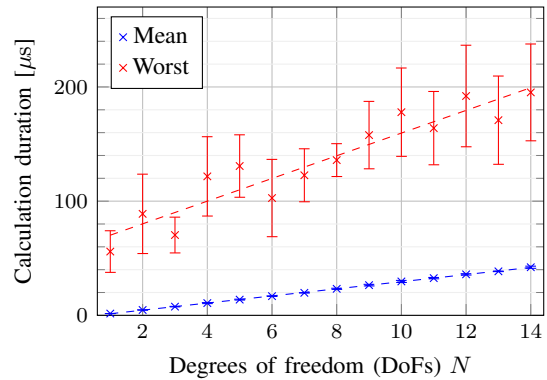
\begin{figure}[ht]
    \centering
\pgfplotstableread{calculation-duration.txt}{\times}
\begin{tikzpicture}
\begin{axis}[
    xmin = 0.6, xmax = 14.4,
    ymin = 0, ymax = 270,
    grid = both,
    major grid style = {lightgray},
    minor grid style = {lightgray!25},
    minor y tick num=4,
    width = 0.46\textwidth,
    height = 0.35\textwidth,
    legend pos = north west,
    legend style={font=\small},
    xlabel = {\Acfp{DoF} $N$},
    ylabel = {Calculation duration [\si{\mu s}]}
]

    \addplot+[only marks, blue, mark=x, error bars/.cd, y fixed, y dir=both, y explicit] table [x=dof, y=mean, y error=mean-err] {\times};
    \addplot+[only marks, red, mark=x, error bars/.cd, y fixed, y dir=both, y explicit] table [x=dof, y=worst, y error=worst-err] {\times};
    \legend{Mean, Worst}
    
    \addplot[red, dashed] table[x=dof, y={create col/linear regression={y=worst}}] {\times};
    \addplot[blue, dashed] table[x=dof, y={create col/linear regression={y=mean}}] {\times};
\end{axis}
\end{tikzpicture}
    \caption{Calculation performance depending on the number of \acfp{DoF}.}
    \label{fig:calculation-duration-dof}
\end{figure}

Table~\ref{tab:calculation-duration} shows the mean and worst calculation performance for a robotic system with 7 \acp{DoF}. Fig.~\ref{fig:calculation-duration-dof} shows the calculation duration depending on the number of \acp{DoF}.

\begin{figure*}[t]
    \centering
\begin{tikzpicture}
\begin{axis}[
    width=0.5\textwidth,
    height=0.6\textwidth,
    xmin=0.0, xmax=0.6,
    ymin=-0.01, ymax=0.71,
    grid=major,
    grid style=dashed,
    xlabel={$x$ [m]},
    ylabel={$y$ [m]},
    title={\small Trajectory - Top View},
]

\draw[fill=black!50, fill opacity=0.1] (0, 0.5) rectangle (4, 2.0);

\addplot[thick] table {hrc-data.txt};

\draw[red, dashed] (0.35, 0.25) circle (0.125);
\draw[->, thick, dashed] (0.48, 0.065) -- (0.41, 0.17);

\node[blue] (a) at (0.05, 0.32) {$\times$};
\node[blue, below of=a, node distance=10pt] (albl) {A};

\node[blue] (b) at (0.55, 0.205) {$\times$};
\node[blue, below of=b, node distance=10pt] (blbl) {B};

\node[red] (c) at (0.35, 0.25) {$\times$};
\node[red, below of=c, node distance=10pt] (clbl) {C};

\node[red] (d) at (0.4, 0.6) {$\times$};
\node[red, above of=d, node distance=10pt] (dlbl) {D};

\draw[thick, postaction={decorate}, decoration={markings, mark=at position 0.5 with {\arrow{>}}}] (0.13, 0.62)--(0.05, 0.32);
\draw[thick, postaction={decorate}, decoration={markings, mark=at position 0.5 with {\arrow{>}}}] (0.05, 0.32)--(0.35, 0.25);
\end{axis}

\pgfplotstableread{hrc-x.txt}{\profilea}
\begin{axis}[
    xmin = 0, xmax = 1.667,
    ymin = -1.8, ymax = 1.8,
    xtick distance = 0.2,
    ytick distance = 0.5,
    grid = major,
    minor tick num = 1,
    at={(0.51\linewidth, 0.26\textwidth)},
    major grid style = {lightgray!50},
    width = 0.49\textwidth,
    height = 0.35\textwidth,
    legend style={
        at={(0.5, 1.03)},
        anchor=south,
        legend columns=-1
    },
    ylabel={\ac{DoF} $x$ [a.u.]}
]
	\draw [fill, orange!50, fill opacity=0.1] (-1, 0.1) rectangle (4, 2);
	\draw [fill, teal!50, fill opacity=0.1] (-1, 1.6) rectangle (4, 2);
	\draw [fill, teal!50, fill opacity=0.1] (-1, -1.6) rectangle (4, -2);
	
	\draw [thin, dashed, teal] (-1, 1.6) -- (4, 1.6);
    \draw [thin, dashed, teal] (-1, -1.6) -- (4, -1.6);
    \draw [thin, dashed, orange] (-1, 0.1) -- (4, 0.1);
    \draw [thin, dashed, orange] (-1, -2.0) -- (4, -2.0);

    \addplot[teal, thick] table [x={t}, y={a}] {\profilea};
    \addplot[orange, thick] table [x={t}, y={v}] {\profilea};
    \addplot[blue!80!black, thick] table [x={t}, y={p}] {\profilea};
    \legend{Acceleration, Velocity, Position}
\end{axis}

\pgfplotstableread{hrc-y.txt}{\profileb}
\begin{axis}[
    xmin = 0, xmax = 1.667,
    ymin = -1.6, ymax = 1.6,
    xtick distance = 0.2,
    ytick distance = 0.5,
    grid = major,
    minor tick num = 1,
    at={(0.51\linewidth, -0.03\textwidth)},
    major grid style = {lightgray!50},
    width = 0.49\textwidth,
    height = 0.35\textwidth,
    ylabel={\ac{DoF} $y$ [a.u.]},
    xlabel={Time $t$ [s]}
]

	\draw [fill, orange!50, fill opacity=0.1] (-0.2, -0.05) rectangle (4, -2);
	\draw [fill, teal!50, fill opacity=0.1] (-1, 1.5) rectangle (4, 2);
	\draw [fill, teal!50, fill opacity=0.1] (-1, -1.5) rectangle (4, -2);
	
	\draw [thin, dashed, teal] (-1, 1.5) -- (4, 1.5);
    \draw [thin, dashed, teal] (-1, -1.5) -- (4, -1.5);
    \draw [thin, dashed, orange] (-1, 3.0) -- (4, 3.0);
    \draw [thin, dashed, orange] (-1, -0.05) -- (4, -0.05);

    \addplot[teal, thick] table [x={t}, y={a}] {\profileb};
    \addplot[orange, thick] table [x={t}, y={v}] {\profileb};
    \addplot[blue!80!black, thick] table [x={t}, y={p}] {\profileb};
\end{axis}
\end{tikzpicture}
    \caption{An example application in the field of \acf{HRC}: The robot pick-and-places an object from (A) to (B). However, a worker accidentally enters the range of the robot (dashed) and triggers a safety violation at (C). Ruckig calculates a time-optimal trajectory within one control cycle to a pre-defined state (D) within a safe zone (gray). Notably, the velocity limits towards the human are near zero (due to the safety) and much larger in the opposed direction (limited by the robot dynamics). This results in the desired behavior that the robot first brakes the velocity as fast as possible and then moves to the safe zone.}
    \label{fig:hrc-example}
\end{figure*}
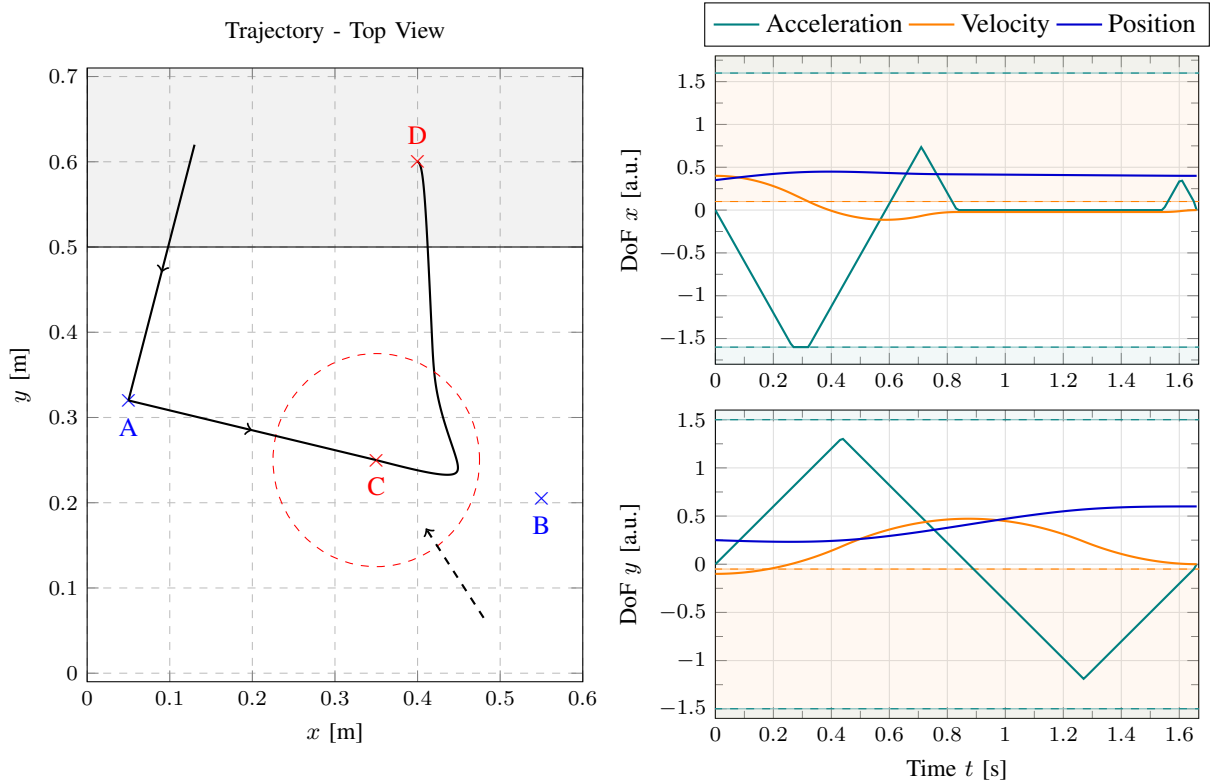

As expected, we find a near linear relationship $O(N)$ between the average performance and the number of \acp{DoF}. Furthermore, we find that \textit{Ruckig} is well suited for control cycles as low as half a millisecond. In fact, $N=82$ is the smallest number of \acp{DoF} that misses the control cycle of \SI{1}{ms} in the worst case on our hardware.

\subsection{Real-world Experiments}

We have integrated \textit{Ruckig} into our controller library \textit{frankx} for the Franka Emika Panda robot arm with \num{7} \acp{DoF}. \textit{Frankx} is available under the MIT license at \url{https://github.com/pantor/frankx} and allows for high-level motion generation. As the Franka robot checks for acceleration discontinuities in its real-time control, a constrained jerk is a hard requirement for \textit{frankx}. The robot has a control cycle time of \SI{1}{ms}. \\

\textit{Ruckig} is agnostic towards the used parametrization; commonly either the joint space or the Cartesian (task) space formulations are used. Furthermore, we use Cartesian space control with 3 translational, 3 rotational \acp{DoF} and a single elbow parameter. Furthermore, we highlight two possible applications.

\subsubsection{Online Reaction to Sensor Input}

A robot requires \ac{OTG} to react to unforeseen sensor input. Fig.~\ref{fig:hrc-example} shows a concrete example within the field of \ac{HRC}. Here, the position of a human worker is monitored. After a detected safety violation, the velocity towards the human should be reduced to a minimum to weaken possible collisions. Moreover, the robot should move away from the human as fast as possible to avoid bruises or potential dangerous contact. \textit{Ruckig} offers a high-level interface for this scenario: The robot moves as fast as possible to a safe target position with zero velocity, limited by a near-zero velocity $v_{max}$ towards the human, and a robot-limited velocity $v_{min}$ in the opposite direction. 

\subsubsection{Offline Trajectory Planning}

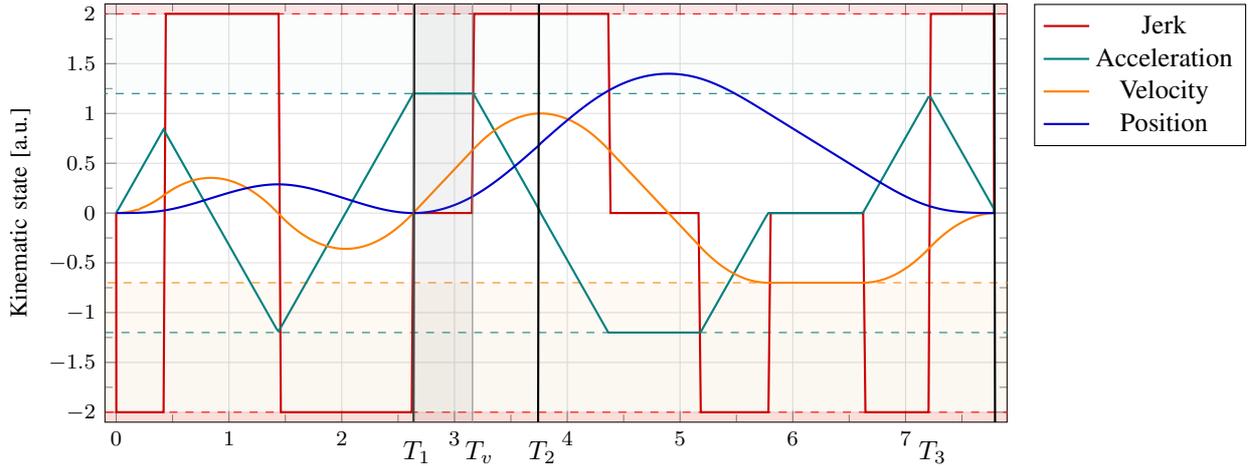
\begin{figure*}
    \centering
\pgfplotstableread{offline-planning.txt}{\profile}
\begin{tikzpicture}
\begin{axis}[
    xmin = -0.1, xmax = 7.9,
    ymin = -2.1, ymax = 2.1,
    xtick distance = 1,
    ytick distance = 0.5,
    grid = major,
    minor tick num = 1,
    major grid style = {lightgray!50},
    width = 0.83\textwidth,
    height = 0.44\textwidth,
    legend pos=outer north east,
    ylabel={Kinematic state [a.u.]}
]

    \draw[teal, fill, fill opacity=0.018, dashed] (-1, 1.2) rectangle (8, 3);
    \draw[teal, fill, fill opacity=0.018, dashed] (-1, -1.2) rectangle (8, -3);
    \draw[orange, fill, fill opacity=0.05, dashed] (-1, -0.7) rectangle (8, -3);
    \draw[red, fill, fill opacity=0.1, dashed] (-1, 2.0) rectangle (8, 3);
    \draw[red, fill, fill opacity=0.1, dashed] (-1, -2.0) rectangle (8, -3);
 
    \addplot[red!80!black, thick] table [x={t}, y={j}] {\profile};
    \addplot[teal, thick] table [x={t}, y={a}] {\profile};
    \addplot[orange, thick] table [x={t}, y={v}] {\profile};
    \addplot[blue!80!black, thick] table [x={t}, y={p}] {\profile};
    \legend{Jerk, Acceleration, Velocity, Position}
    
    \draw[thick, black] (2.635, -3) -- (2.645, 3);
    \draw[thick, black] (3.74, -3) -- (3.75, 3);
    \draw[thick, black] (7.795, -3) -- (7.785, 3);
    
    \draw[very thin, gray, fill, fill opacity=0.1] (2.635, -3) rectangle (3.16, 3);
\end{axis}

\node[] (t1) at (4.1, -0.4) {$T_1$};
\node[] (tv) at (4.92, -0.4) {$T_v$};
\node[] (t2) at (5.75, -0.4) {$T_2$};
\node[] (t3) at (10.88, -0.4) {$T_3$};

\end{tikzpicture}
    \caption{An example for offline planning of the following task: An object placed at $p=0$ should be accelerated to a target velocity (without jerk constraints) in the shortest time possible by a robot (with jerk constraint). The robot starts from rest ($T_0$). As no impact should occur (otherwise leading to an acceleration violation), the contact between object and robot (gray) should happen at zero velocity but with maximum acceleration ($T_1$). After the  object reaches its velocity ($T_v$), the robot brakes and reaches its maximum velocity ($T_2$). Then, the robot moves back to its origin ($T_3$) with reduced velocity.}
    \label{fig:offline-planning-example}
\end{figure*}

Ruckig allows for trajectory planning by following a list of successive waypoints. In particular, waypoints with zero target velocity and acceleration correspond the a piecewise path without blending. Given following waypoints,
\begin{table}[h]
	\centering
	\begin{tabular}{c|c|c|c}
	\hline
	& Position $[m]$ & Velocity $[m/s]$ & Acceleration $[m/s^2]$ \\
	\hline
	0 & $0$ & $0$ & $0$ \\
	1 & $0$ & $0$ & $1.2$ \\
	2 & $0.68$ & $1.0$ & $0$ \\
	3 & $0$ & $0$ & $0$ \\
	\hline
	\end{tabular}
	\label{tab:offline-waypoint-list}
\end{table}
we highlight the use of a non-zero acceleration target in a dynamic task: An object (without any jerk constraint) should be accelerated as fast as possible by a jerk-limited robot. Therefore, no impact between robot and object should occur, as this would lead to an acceleration violation. Then, the robot should only be in contact with the object at its maximum acceleration. After reaching the target velocity of the object, the robot smoothly decelerates and moves back to its initial position. Fig.~\ref{fig:offline-planning-example} illustrates the resulting trajectory.

\section{Discussion and Outlook}

We presented \textit{Ruckig}, an \acf{OTG} algorithm that is able to handle non-zero target accelerations. As \textit{Ruckig} considers third-order constraints (for velocity, acceleration, and jerk), the complete kinematic state can be specified for waypoint-based trajectories. The proposed algorithm is real-time capable and of high performance regarding control cycles as low as \SI{1}{ms}. Our implementation is available as an open-source C++ library. \\

In comparison to related work, \textit{Ruckig} expands the capabilities of the proprietary Reflexxes Type IV library \cite{kroger2011opening}. Reflexxes uses \textit{decision trees} to find matching profiles as well as the blocked intervals, and calculates the numerical profile afterwards. With over \num{4760} unique nodes, these decision trees cause great complexity \cite{kroger2009online}. In contrast, Ruckig calculates all valid extremal profiles first and derives matching profiles and blocked intervals afterwards. The proposed algorithm is significantly simpler without decision trees: The relevant code-base of Ruckig has around \num{2800} lines of code in comparison to \num{23000} lines for Reflexxes Type IV. To our own surprise, we find that the mean calculation performance of Ruckig is around twice as fast as Reflexxes. This is probably due to implementational details and better optimizations. In comparison to opt$\_$control \cite{beul2016analytical}, Ruckig is able to handle blocked intervals for time synchronization and therefore \textit{guarantees} a solution for arbitrary input states. Moreover, Ruckig is an order of magnitude faster and real-time capable. In contrast to other related work within the field of \ac{OTG}, Ruckig supports a complete initial and target state \cite{broquere2008soft, haschke2008line}, time-optimality \cite{ahn2004arbitrary, wang2020research}, and multiple \acp{DoF} \cite{macfarlane2003jerk}. \\

Furthermore, Kröger et al. showed that \ac{OTG} can be applied to motion planning using intermediate waypoints \cite{kroger2009online}. We argue that Ruckig is much more suitable for this application: By defining waypoints with non-zero target accelerations, curves, circles, or splines can be better approximated, resulting in smoother motions. In contrast, waypoints with zero velocity result in plateaus of constant velocity. In the future, we want to further investigate this application to bridge the gap between \ac{OTG} and path-following time-parametrization.

\bibliographystyle{plainnat}
\bibliography{references}

\end{document}